\documentclass{article}

\usepackage[preprint]{neurips_2026}

\usepackage[utf8]{inputenc}
\usepackage[T1]{fontenc}
\usepackage{hyperref}
\usepackage{url}
\usepackage{booktabs}
\usepackage{amsfonts}
\usepackage{nicefrac}
\usepackage{microtype}
\usepackage{xcolor}
\usepackage{graphicx}
\usepackage{amsmath}
\usepackage{comment}

\renewcommand{\cite}[1]{\citep{#1}}

\title{Inpainting physics: self-supervised learning for context-driven fluid simulation}

\author{
  \textbf{Jonas Weidner}$^{1,2}$,
  \textbf{Yeray Martin-Ruisanchez}$^{1}$,
  \textbf{Daniel Rueckert}$^{1,2,3}$, \\
  \textbf{Benedikt Wiestler}$^{1,2,*}$,
  \textbf{Julian Suk}$^{1,2,*}$ \\[0.5em]
  {\normalfont $^1$ Technical University of Munich (TUM)} \\
  {\normalfont $^2$ Munich Center for Machine Learning (MCML)} \\
  {\normalfont $^3$ Imperial College London} \\
  {\normalfont $^*$ Shared senior authorship} \\
  {\normalfont\texttt{j.weidner@tum.de}
  }
}

\begin{document}

\maketitle
\begin{abstract}

Neural surrogate models for computational fluid dynamics (CFD) are typically trained as forward operators that map explicit problem specifications, such as geometry and boundary conditions, to solution fields. This ties the model to the conditioning variables seen during training and limits reuse under boundary-condition shifts or local geometry changes. We propose to reformulate steady CFD inference as an inpainting problem: instead of training on explicit boundary conditions, we learn a self-supervised prior over velocity fields and impose boundary constraints only during inference by fixing known regions such as inlet, outlet
or unchanged regions from previous simulations. To scale this idea to large 3D meshes, we introduce a local neighbourhood tokeniser that represents high-resolution velocity fields as compact spatial latent tokens and train latent flow-matching and masked-autoencoder models on these tokens. On simulations of intracranial aneurysm flow, our method reconstructs full velocity fields from sparse boundary context, outperforms supervised neural surrogates under boundary-condition and dataset shift and enables local geometry editing by reusing unchanged simulation context.
These results suggest that viewing CFD inference as context-conditioned inpainting can turn neural surrogates from task-specific predictors into reusable flow priors. Code and demo are available at \href{https://inpainting-physics.com}{inpainting-physics.com}.

\end{abstract}

\textbf{TL;DR} Train a self-supervised latent model on CFD velocity fields, impose boundary conditions at inference, predict the flow by inpainting, generalise more flexibly to unseen geometries and flow speeds.

\section{Introduction}

Computational fluid dynamics (CFD) is a central tool in, e.g., automotive engineering, climate science and biomedical modelling, but high-fidelity numerical simulations remain expensive~\cite{ashton2024drivaerml,bodnar2025foundation,nannini2024automated}.
Each new geometry, boundary condition or design variant typically requires solving a large discretised system from scratch.
Neural surrogates aim to reduce this cost by learning an approximate map from problem specification to quantities of interest.

Neural surrogates for CFD are typically trained as supervised neural operators: given a defined geometry, physical parameters and boundary conditions, they directly regress the corresponding velocity or pressure field~\cite{suk2024lab,luo2025transolver++,alkin2025ab}.
This formulation is powerful in-distribution, but strongly couples the learned model to the conditioning variables and data-generating process used during training.
As a result, extrapolation to new boundary conditions, geometries or solver settings remains a major obstacle.

We begin by observing that laminar fluid flow correlates highly with its spatial context.
Due to viscosity effects and momentum, the union of upstream and downstream dynamics is an excellent prior for local flow.\footnote{The totality of such local interactions are the basis of continuum-based fluid simulation.}
Despite this, supervised neural operators do not explicitly leverage spatial context when producing flow estimates.
We argue that including spatial context could greatly improve out-of-distribution performance in neural surrogates.

Inspired by the success of
generative modelling in computer vision~\cite{rombach2022high},
we formulate (steady) CFD emulation not as deterministic regression but as (probabilistic) generation.
In other words, we recast the supervised learning of flow states based on their underlying Dirichlet boundary conditions to self-supervised, \emph{unconditional} learning in conjunction with \emph{conditioning at inference time}.
Our goal is to learn a reusable prior over all flow states and impose Dirichlet boundary conditions only at inference time via \emph{inpainting}.
The generative model can then leverage upstream and downstream observations and infer dynamical parameters implicitly from the context.\footnote{We remark that the inpainting formulation smoothly transitions into the regression formulation if we increase the size of the inpainting region (until the Dirichlet boundary).}

This separation of representation learning and context-driven inference changes how neural surrogates can be used.
A pre-trained model can be queried with different conditioning patterns: sparse inlet-outlet constraints (similar to the supervised setting) as well as larger measured or otherwise known spatial context.
Furthermore, the inpainting formulation enables local editing, i.e., adaptation of a given simulation to small geometric perturbations of the flow domain.

\paragraph{Contributions.}
We (i) propose to shift the modelling objective and view CFD emulation as context-conditioned inpainting rather than supervised neural operator modelling, allowing conditioning through observed regions, (ii) show that this approach has favourable generalisation properties for changing boundary conditions and shape distribution, and (iii) propose local editing as a preliminary downstream task tailored to generative models.

\begin{figure}[t]
  \centering
  \includegraphics[width=\textwidth]{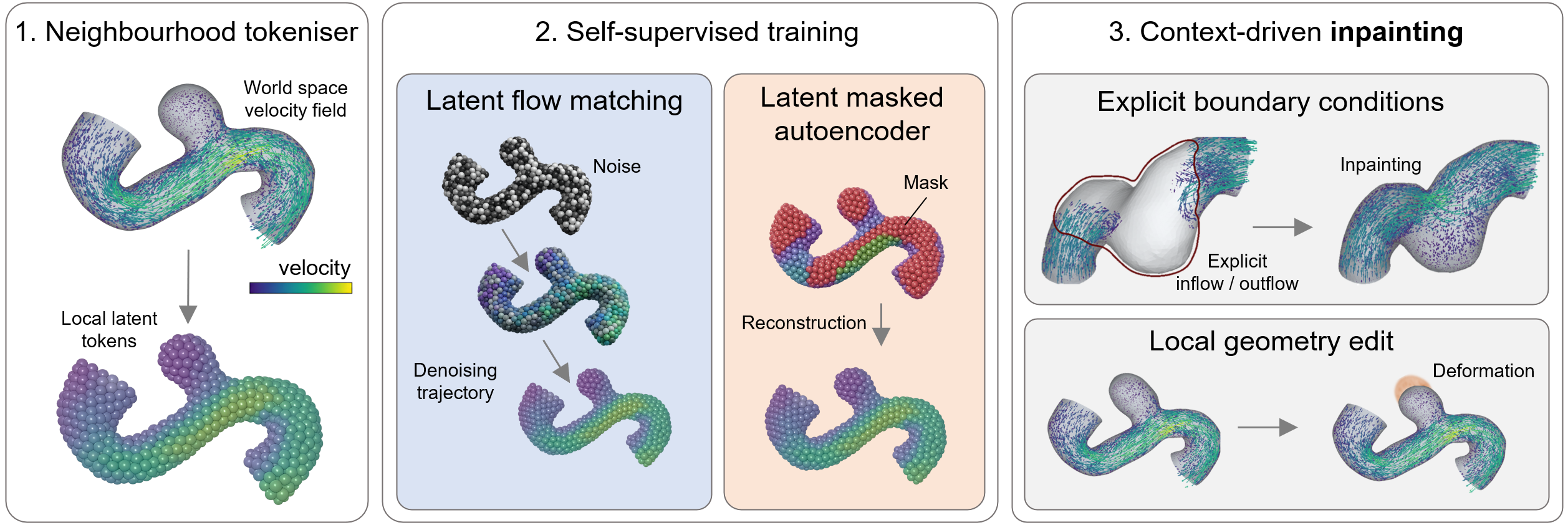}
  \caption{\textbf{Inpainting physics.} 
(1) We tokenise raw velocity fields into local ball-shaped latent representations. 
(2) We train a self-supervised model on these tokenised velocity fields using latent flow matching or a masked autoencoder. 
(3) At inference, known regions (e.g., inflow and outflow, measurements, or unchanged parts of a previous simulation) are fixed as context and the rest is inpainted, enabling generalisation to unseen geometries and flow conditions, preserving local flow structure.}
  \label{fig:overview}
\end{figure}
\section{Related work}
Neural surrogates for science and engineering have recently been gaining traction.
Their underlying idea is to learn a mapping from the problem parameters to the quantities of interest in order to approximate high-fidelity numerical simulation in certain regimes where data is abundant.
Proofs of concept exist, e.g., for weather forecasting~\cite{bodnar2025foundation}, automotive aerodynamics~\cite{alkin2025ab}, high-energy physics~\cite{spinner2024lorentz} and cardiovascular flow~\cite{suk2024lab}.

\paragraph{Neural operators.}
Popularised by architectures like DeepONet~\cite{lu2021learning} and Fourier Neural Operator~\cite{li2021fourier}, ``neural operators" has become an umbrella term for models that map between continuous functions.
In the context of neural surrogates, these models are typically set up as forward operators, mapping geometric and physical inputs to the solution field~\cite{luo2025transolver++,alkin2025ab,pepe2025fengbo,holzschuh2025pde,wen2026geometry,holzschuh2026p3d}.
Conceptually, these approaches are designed for regular grids
or point clouds
extracted from 3D surface or volume meshes.
While grid-based methods lend themselves to Euclidean domains where isotropically high resolution is required, e.g. freestream flow within a volume element, point cloud methods can efficiently model irregular structures and their interaction with the medium, e.g., blood flow in human vasculature.

\paragraph{Generative models.}
Due to their efficacy in domains like image processing and computer vision, generative models are gradually incorporated into neural surrogates.
Diffusion models have been proposed for temporal forecasting of solutions to partial differential equations (PDE) where the main difficulty lies in achieving stable rollouts~\cite{sun2023unifying,lippe2023pde,shysheya2024conditional,rozet2026lost}.
Notably, \citet{rozet2026lost} highlight the efficacy of encoding the data into a unified latent space in which system dynamics are generated.
We adopt latent space modelling as a design choice in our approach.

Instead of autoregressively evolving an initial state over time, \citet{lienen2024zero} propose to learn the manifold of possible flow states in a certain (turbulent) regime constituted by the training data.
In contrast to our approach, \citet{lienen2024zero} impose Dirichlet boundary condition already during training and focus on directly modelling the distribution of turbulence rather than emulating a numerical solver.
\citet{lino2025learning} propose to generate converged flow states via graph-based latent diffusion conditioned on parameters of the dynamics, such as Reynolds number.
While this more closely emulates a numerical solver, \citet{lino2025learning} also impose Dirichlet boundary conditions during training which means their model is tied to these parameters, different from our context-driven approach.
\citet{du2024conditional} propose an unconditional approach to temporal turbulence modelling using a latent space learned via conditional neural fields.
In contrast to their approach, we include variations in mass flow and domain geometry in our framework while \citet{du2024conditional} focus on generating complex turbulence patterns within a fixed setting.

Besides diffusion (or flow matching) models, generative adversarial networks (GAN) have been explored in the context of fluid modelling~\cite{buzzicotti2021reconstruction,drygala2022generative}.
In addition to these two approaches, inference-time flow estimation via inpainting has been previously studied~\cite{buzzicotti2021reconstruction,du2024conditional}.
However, those works are limited to grid-based data (i.e., images).
In contrast, we use inpainting as the conditioning mechanism itself, replacing explicit boundary-condition parameters, and lay out a framework for inpainting on point clouds from 3D volume meshes, which is the dominant data modality for state-of-the-art science and engineering problems.
Additionally, our work is unique in identifying the utility of masked autoencoders (MAE)~\cite{he2022masked} as a generative backbone.

Recently, \citet{garnier2025meshmask} proposed masked graph neural networks for pre-training neural surrogates on (2D and) 3D CFD problems.
Their approach differs from ours in that masking is integrated with their autoencoder: \citet{garnier2025meshmask} mask inputs before passing them to the encoder which performs message passing within the resulting context while local completion of masked regions is done during the decoding process.
In contrast, we mask and complete locality-preserving ``latent" tokens.

\section{Methods}
\subsection{Datasets}
Similar approaches to ours commonly study automotive aerodynamics in the context of neural surrogates~\cite{luo2025transolver++,alkin2025ab}, characterised by freestream flow around an object.
In order to clearly demonstrate the mechanism of our approach, we seek an application in which (laminar) dynamics are dominated by upstream velocity rather than domain geometry.
For our application, arterial blood flow is particularly well-suited, with two datasets of intracranial aneurysm fluid simulations~\cite{li2025aneumo,ding2025aneug} available.
These
large-scale public datasets allow for benchmarking and out-of-distribution testing, especially since their samples vary in mass flow and shape.

Specifically, we train and evaluate our method on the Aneumo dataset \cite{li2025aneumo}. The data contains multiple slightly adapted aneurysm geometries derived from real patients. These samples are, therefore, an ideal test bed for studying both full-velocity field generation and local editing around the aneurysms. The dataset contains 85,280 samples of steady-state fluid simulations across 8 total mass flows, the mass that flows through the artery per given time. Units for velocities are [m/s] and mass flow [g/s] if not stated otherwise. We train our model on 4 different mass flows (2, 2.5, 3, 3.5) and keep (1, 1.5, 3.75, 4) for testing. For all models, we center the simulation at the centre of mass and add a small translation augmentation. 
To prevent data leakage, we split the data sequentially (test 1 - 1,093, evaluating on a uniformly sampled subset of 100 due to L-FM integration cost; validation 1,094 - 2,158; training 2,159 - 10,660), as clusters of adjacent samples have similar geometry. 

As an additional test, we evaluate our methods on the transient dataset AneuG~\cite{ding2025aneug}, averaging over all timesteps to obtain a stationary estimate. The mass flows per artery are not provided; thus, we estimate a constant scaling factor of 86.97 based on the mean velocity fitted to Aneumo. AneuG uses non-Newtonian Carreau–Yasuda blood, while Aneumo uses a Newtonian fluid. Reference solutions further come from different solver stacks.
We use AneuG strictly as a held-out test set to assess model generalisation, with no cases seen during training or validation.

\subsection{Model}

We formulate CFD prediction as conditional completion under a self-supervised velocity field prior. The model is trained on complete velocity fields $v$ without explicit boundary-condition labels, such as mass flow, inlet profiles or outlet constraints. At inference time, known solution regions are fixed as spatial context and the model reconstructs the unknown (masked) region, estimating $v_{\mathcal U} \mid v_{\mathcal K}$, where $\mathcal K$ denotes known regions such as inlet, outlet, measured, or unchanged simulation regions, and $\mathcal U$ denotes the region to be inpainted.

Our method has three stages, illustrated in Figure~\ref{fig:overview}: a local neighbourhood tokeniser compresses the unstructured velocity field into spatial latent tokens (Sec.\ref{sec:nt}); a self-supervised generative model, either latent flow matching or a latent masked autoencoder, is trained on these tokens (Sec.\ref{sec:ssm}); and inference enforces known context by fixing the corresponding tokens while inpainting the missing region (Sec.~\ref{sec:inference}).

\subsubsection{Neighbourhood tokeniser}
\label{sec:nt}
The velocity field is represented on an unstructured three-dimensional point cloud. Each point $i$ has position $x_i \in \mathbb{R}^3$ and velocity $v_i \in \mathbb{R}^3$. We additionally provide each point with its Euclidean distance to the closest physical wall, $d_i \in \mathbb{R}$. Direct transformer modelling of the full point cloud is infeasible because each simulation contains approximately $200\mathrm{k}$ points. Moreover, our inpainting formulation requires spatially localised tokens: when a boundary or edited region is fixed, the representation should be constrained to this specific location. We therefore introduce a local neighbourhood tokeniser that maps the dense velocity field to a set of compact latent tokens.

We first select $P=2500$ neighbourhood centres $x_j^p$ using farthest point sampling~\citep{eldar1997farthest}. Around each centre, we collect the $512$ nearest points. Each point in a neighbourhood is represented by its local coordinate relative to the neighbourhood centre normalised by the radius $r$ (distance to furthest point from the centre), its wall distance, and its velocity:
\begin{equation}
    \left[\tilde{x}_i, d_i, v_i\right] \in \mathbb{R}^7,
    \qquad
    \tilde{x}_i = (x_i - x_j^p)/r.
\end{equation}
A PointNet-style encoder~\citep{qi2017pointnet} with shared MLP layers maps the $512$ point features in each neighbourhood to pointwise features, followed by max pooling to obtain one latent token $z_j \in \mathbb{R}^{256}$. The decoder reconstructs local velocities by broadcasting $z_j$ to query points and concatenating the local geometric descriptor $(\tilde{x}_i, d_i)$. Each neighbourhood is decoded independently. When a point belongs to multiple decoded neighbourhoods, we combine predictions using inverse-distance weighting in world space. Details are provided in Appendix~\ref{Appendix:NT}. This produces a full-resolution velocity field while preserving the locality needed for inpainting.
The tokeniser is trained independently with a mean-squared-error reconstruction objective. Its role is not to solve the CFD task, but to provide a compact, smooth, and spatially localised latent space in which global generative models can operate efficiently.

\subsubsection{Self-supervised model}
\label{sec:ssm}

After tokenisation, each simulation is represented by a set of latent tokens $Z$. The self-supervised model is trained on these latent fields. It receives token positions and wall distances to learn spatially plausible velocity field representations, but it does not receive total mass flow, inlet velocity, outlet conditions, or a global geometry descriptor. We therefore call the model unconditional with respect to boundary conditions and simulation parameters.

We evaluate two complementary self-supervised backbones: latent flow matching and a latent masked autoencoder.

\paragraph{Latent flow matching.}

The latent flow-matching model learns a continuous transport from a simple noise distribution to the distribution of latent velocity fields. For each latent field $Z$, we sample Gaussian noise $\epsilon \sim \mathcal{N}(0, I)$ with the same shape as the latent tokens and time $t \sim \mathrm{LogitNormal}(0, 1)$. We define the linear interpolant
\begin{equation}
    z_t^p = (1-t)\epsilon + t z^p.
\end{equation}
The model $f_\theta$ is trained to predict the velocity:
\begin{equation}
    \mathcal{L}_{\mathrm{FM}}
    =
    \mathbb{E}_{t,\epsilon,z^p}
    \left[
    \left\|
    f_\theta(z_t^p, t, x^p, d^p)
    -
    (z^p - \epsilon)
    \right\|_2^2
    \right].
\end{equation}
The backbone is a DiT-style transformer~\cite{peebles2023scalable} with adaLN-Zero conditioning on $t$. Spatial positions and wall distances are encoded with sinusoidal embeddings. At inference time, unknown tokens are initialised with noise or zeros and integrated toward the data distribution (Sec.~\ref {sec:inference}), while known tokens are fixed to their encoded values throughout the integration.

\paragraph{Latent masked autoencoder.}

The latent masked autoencoder is trained directly for conditional completion. Given a latent field, we randomly mask a subset of token indices $\mathcal M$. The encoder receives only the visible tokens, together with their spatial and wall-distance embeddings. The decoder receives encoded visible tokens and learned mask tokens at the missing positions, and predicts the original latent tokens at the masked locations. The training objective is
\begin{equation}
    \mathcal{L}_{\mathrm{MAE}}
    =
    \frac{1}{|\mathcal M|}
    \sum_{j \in \mathcal M}
    \left\|
    \hat{z}_j - z_j
    \right\|_2^2.
\end{equation}
This objective closely matches our inference problem: given known regions of a velocity field, predict the missing region. Unlike the flow-matching model, the masked autoencoder does not require iterative sampling at inference time, making it a direct and efficient completion model. We implement a Transformer-based MAE \cite{he2022masked}.

\subsubsection{Explicit boundary inpainting.}
\label{sec:inference}

At inference time, boundary conditions and other known information are imposed by explicitly fixing the corresponding spatial tokens. We encode the known velocity values with the neighbourhood tokeniser and keep those latent tokens fixed. The self-supervised model then predicts the missing tokens in $\mathcal{U}$, after which the NT decoder reconstructs the full-resolution velocity field.

This procedure converts boundary-condition enforcement into an inpainting operation. Importantly, the same pretrained model can be used with different choices of $\mathcal K$ (Figure~\ref{fig:overview}). For full-field prediction, $\mathcal K$ may consist only of sparse inlet and outlet regions. For local geometry editing, $\mathcal K$ can include most of the original simulation, while only the edited region is masked and regenerated.

We compare the following inference strategies:
\textbf{L-MAE:} The masked region is predicted in one forward pass using the latent masked autoencoder.
\textbf{L-MAE-Iterative:} The masked region is filled progressively from the boundary of the known region. We iterate over 5 steps and separate the mask into 5 equal regions.
\textbf{L-FM-Euler:} We integrate over the FM output starting from random noise using Euler integration with 20 steps while keeping the explicit boundary fix. 
\textbf{L-FM-Euler-Zero:} We integrate in the same way as FM-Euler, but we start at zero instead of Gaussian noise. 
\textbf{L-FM-Physics:} We guide the integration steps by conditioning on zero divergence as a hard side constraint~\citep{utkarsh2026physics}, which enforces an incompressible flow. Each FM step is extrapolated into world space, projected onto a solution that satisfies the side constraints, and interpolated back into the partial noise space. This forces the integration along physically plausible paths.

The key point is that boundary conditions are never learned as a fixed input format. They are supplied as observed parts of the solution field at inference time. This allows the model to behave like a reusable CFD prior rather than a task-specific forward surrogate.

\subsection{Baselines and metrics.}
We implement a state-of-the-art neural surrogate model as a supervised baseline. They receive the positions as input and output the velocity conditioned on the mass flow.
Therefore, we adopt the anchor-branched universal physics transformer AB-UPT~\citep{alkin2025ab}, which has demonstrated strong performance on large point clouds as an attention-based supervised model (supervised-Att). Additionally, we add a message passing (supervised-MP) model based on PointNet++\cite{qi2017pointnet++}. As our task is defined solely in terms of the velocity field, we exclude the surface and geometry branches and the pressure prediction. We provide the exact same input as for our models $x$ and $d$, but additionally also the Euclidean distance to the inlet $d^{\text{in}}$ for orientation information and the total mass flow $m$ conditioning.
Further, we add the naive baselines: mean-inpainting, which sets all values to the mean, and interpolation, which interpolates based on distance to the nearest non-masked values.

We evaluate the predicted velocity fields along two complementary axes. First, we measure whether \emph{the predicted velocity vectors match the reference CFD solution} at the original point-cloud locations. For this, we report normalised mean squared error (nMSE), which captures errors in velocity magnitude and components, and cosine similarity (CosSim), which measures directional agreement independently of scale. Second, we assess \emph{local physical consistency}. Since the velocity field represents an incompressible flow, we estimate the normalised divergence from local neighbourhoods as a scale-invariant measure of violation of the continuity constraint. In addition, we compare the predicted and reference vorticity fields to evaluate whether the model captures rotational flow structures. Together, these metrics distinguish numerical agreement with the reference solution from physically relevant properties of the reconstructed field. Exact definitions and implementation details are provided in Appendix~\ref{appendix:metrics}.

\section{Experiments}

We evaluate three central questions. First, we test whether a locally trained tokeniser can faithfully represent large velocity fields. Second, we investigate whether boundary-conditioned inpainting can recover full velocity fields under both in-distribution and out-of-distribution conditions. Third, we analyse whether inpainting can exploit partial context to adapt velocity fields under local geometry edits.

\paragraph{Neighbourhood tokens accurately represent velocity fields.}
\begin{figure}[th]
  \centering
  \includegraphics[width=\textwidth]{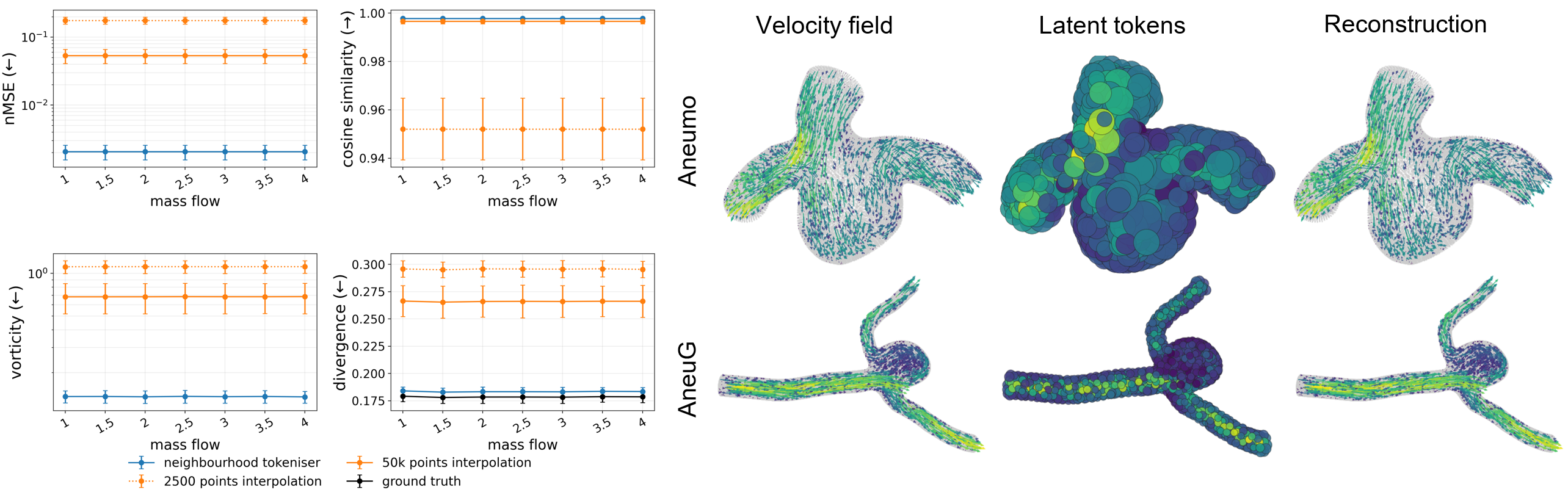}
  \caption{\textbf{The neighbourhood tokeniser demonstrates low reconstruction error over all mass flows.} We compare our tokeniser using 2500 tokens with naive baselines of random downsampling and re-interpolation. Additionally, we show examples of the latent ball representation and the reconstruction on both datasets.
  }
  \label{fig:ball_AE}
\end{figure}

Our inpainting formulation requires a representation that is both compact and spatially local: direct transformer modelling of the full point cloud is infeasible for simulations with approximately $200\mathrm{k}$ points due to the quadratic scaling of self-attention, and global pooling would destroy the locality needed to fix inlet, outlet, measurement, or unchanged simulation regions at inference time.

As shown in Figure~\ref{fig:ball_AE}, the neighbourhood tokeniser compresses each velocity field to $2500$ local latent tokens while maintaining low reconstruction error across mass flows - even when they are out-of-distribution. Compared with random downsampling and interpolation, the learned local representation reconstructs velocity magnitude and direction substantially more accurately. Qualitatively, it preserves the main flow structures on both Aneumo and the held-out AneuG dataset; on AneuG, the reconstruction nMSE is $(0.21 \pm 0.01)\%$. This suggests that \emph{the tokeniser accurately captures local geometric and velocity structure in a way that transfers beyond the training geometries}. Since all subsequent models operate in this latent space, the reconstruction quality of the tokeniser defines an approximate lower bound on the error achievable by the inpainting models.

\paragraph{Explicit boundary-conditioned inpainting predicts velocity fields.}

We next test whether a self-supervised model trained without explicit boundary-condition labels can recover full velocity fields when boundary information is provided at inference time. We evaluate on Aneumo at mass flow $m=3$, which lies inside the training range for all models. The mask fraction denotes the portion of the domain to be inpainted: at $99\%$, essentially only inlet and outlet velocities are fixed, while smaller masks (down to $5\%$) provide more context.

\begin{figure}[ht]
  \centering
  \includegraphics[width=\textwidth]{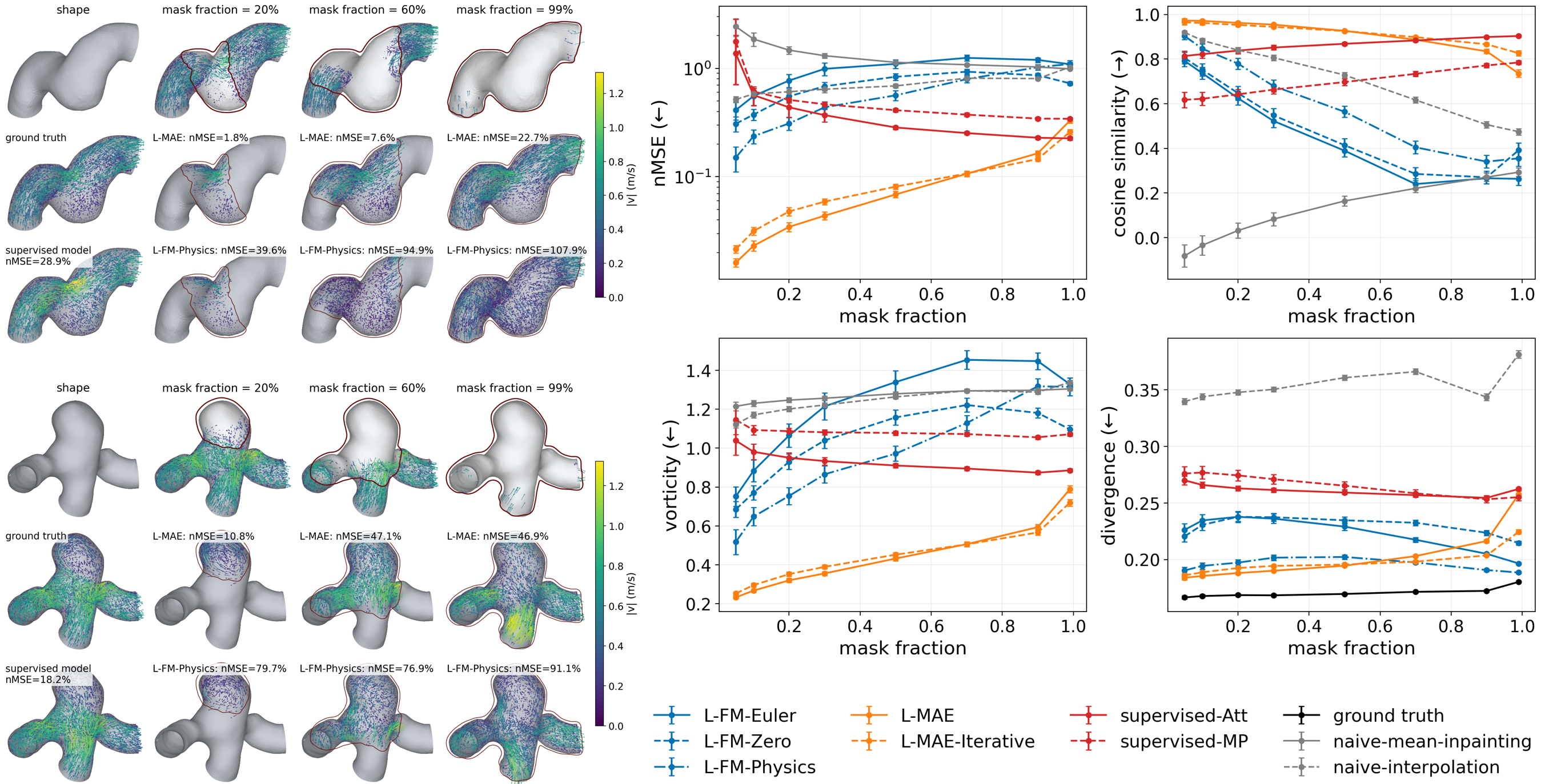}
  \caption{\textbf{Supervised models perform best on the forward prediction on in-distribution tasks. L-MAE and L-FM perform better with additional context.} We compare different supervised and self-supervised models on the prediction of velocity fields provided with varying amounts of context (higher masking fraction means less context). The maximum masking ratio is 99\%.
  Metrics for the supervised models are computed within the mask region.
  }
  \label{fig:mask_frac}
\end{figure}

This setting favours the supervised neural surrogates, which are trained directly for forward prediction and receive the mass flow parameter explicitly. As shown in Figure~\ref{fig:mask_frac}, they therefore perform strongly under sparse (mask fraction $99\%$) in-distribution conditioning. The inpainting models show a different behaviour: their performance improves as more field context is provided. Latent flow matching struggles with very large masks but improves with additional context, and, as expected, physics guidance reduces divergence and vorticity. Further inverse-modelling choices, such as soft boundaries, iterative inpainting, noising the known tokens to the current time, or initialising with the L-MAE prediction, can further improve L-FM (Appendix Figure~\ref{fig:mask_frac_FM}). The latent masked autoencoder is best aligned with the inference task: although slightly weaker than the supervised baseline at the sparsest inlet-outlet setting, it performs best across most metrics once additional context is available. Once about 10\% of the domain is given as context (mask fraction $\leq 90\%$), L-MAE outperforms both supervised surrogates in nMSE, vorticity and divergence.
In other words, context improves localised completion for the former while the latter shine under spatially averaged metrics.
Iterative MAE helps for large masks, but can accumulate errors for smaller ones. Overall, these results show that \emph{boundary-conditioned inpainting can recover full velocity fields from explicit solution contexts, and that its advantage grows as additional known regions become available}.

\paragraph{Inpainting improves out-of-distribution generalisation under boundary-condition and dataset shifts.}

A central motivation for our formulation is robustness to boundary-condition shift. Forward surrogates must extrapolate in an explicit conditioning space, such as total mass flow. Inpainting instead represents new boundary conditions directly as observed velocity values, so the model solves a completion problem in the same field space used during self-supervised training, even though it hasn't seen these specific boundary conditions.

\begin{figure}[ht]
  \centering
  \includegraphics[width=\textwidth]{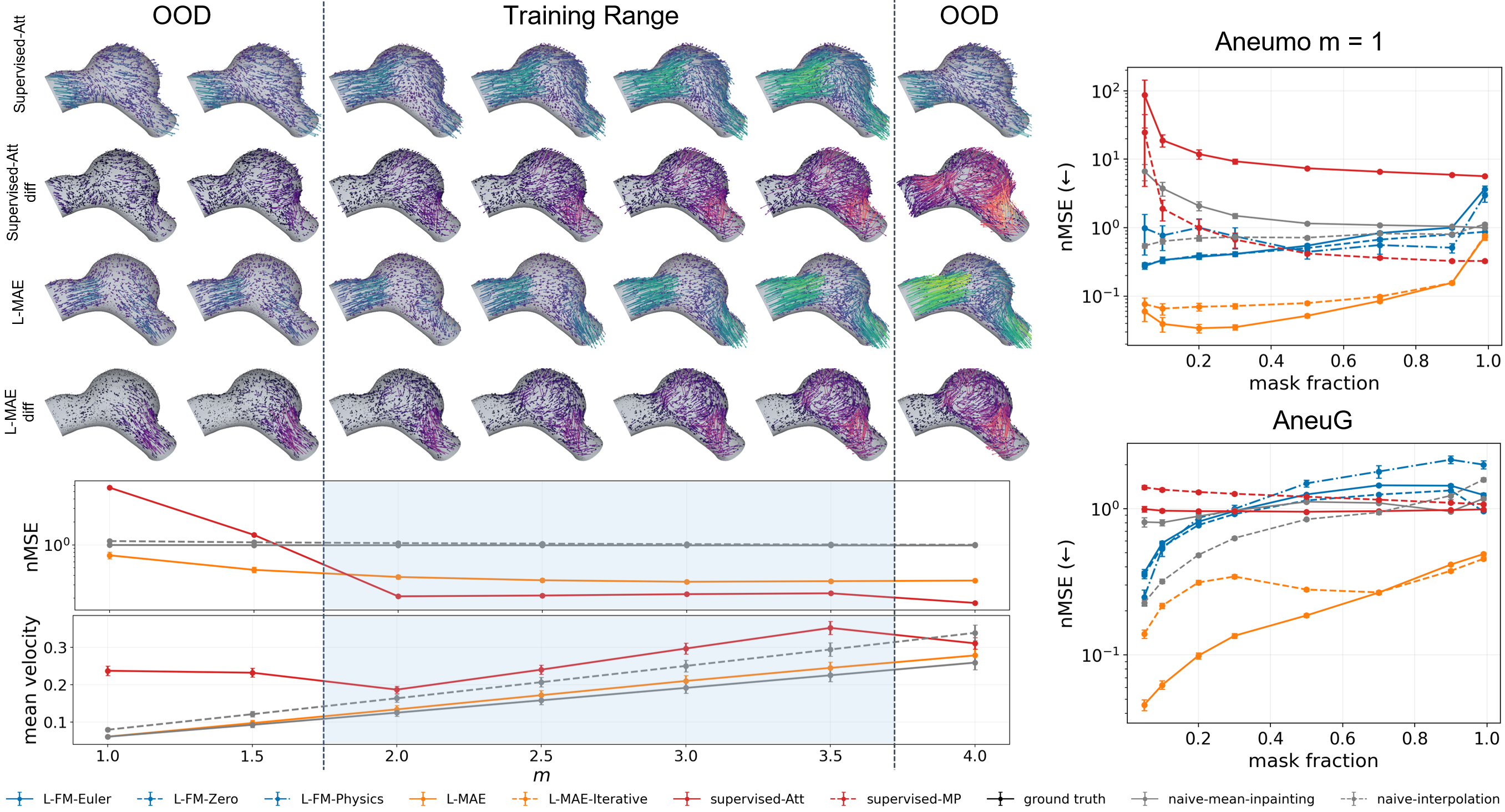}
  \caption{\textbf{The best supervised model (red) fails out-of-distribution (ood), while the L-MAE (orange) provides solid results, especially with context.} Left, we show the performance over all mass flows at a masking ratio of 99\%. Supervised-Att fails to extrapolate the expected linear mean velocity scaling OOD. Right, we provide the nMSE for varied contexts at $m = 1$ and on the external AneuG test set. The maximum masking ratio is 99\%.}
  \label{fig:surrogate_mass_flow}
\end{figure}

Figure~\ref{fig:surrogate_mass_flow} illustrates the limitation of explicit conditioning. Both supervised baselines are trained on mass flows between $2$ and $3.5$ and evaluated from $1$ to $4$. They remain accurate within the training range but collapse for out-of-distribution mass flows, especially for $m=1$. In the inpainting formulation, such cases are not unseen scalar inputs but observed low-velocity boundary regions to be completed; note, however, that our model was still trained only on velocity fields from mass flows between $2$ and $3.5$.

We further test dataset shift on the external AneuG dataset (Appendix Figure~\ref{fig:mask_frac_AneuG}), where geometries, solver stack, mesh resolution, and physical assumptions differ from Aneumo. Under this shift, the supervised forward surrogates degrade strongly and no longer consistently outperform even simple baselines. Among the learned inpainting methods, the latent MAE is the most robust and is the only model that consistently improves over the naive baselines across mask fractions. These results support a main premise of our approach: \emph{conditioning through observed regions of the solution field can be more robust than relying solely on explicit training-time boundary-condition parameters}.

\paragraph{Local geometry editing benefits from reusable global context.}

Local geometry editing highlights another key advantage of inpainting. When only a small part of the domain changes, most of an existing high-fidelity simulation remains valid. A forward surrogate cannot exploit this information and must predict the full edited field from scratch, whereas our model fixes the unchanged region and only infers the modified part.

\begin{figure}[t]
  \centering
  \includegraphics[width=\textwidth]{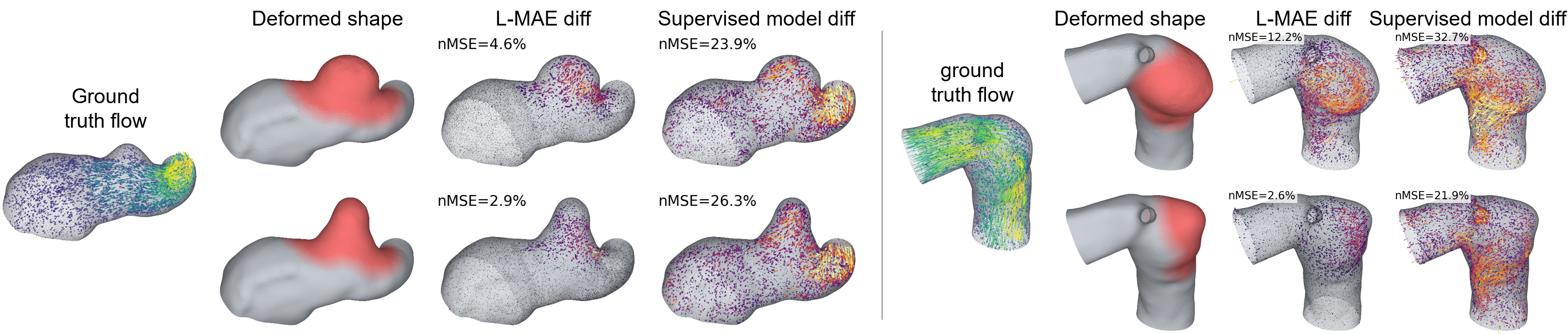}
  \caption{\textbf{Inpainting local geometry edits benefits from global context.} We locally deform two example geometries by modelling the growth of an aneurysm. By generously masking the area around it and conditioning our inpainting approach on the original simulation, we achieve superior results compared to neural surrogates that simulate the full geometry from scratch for every edit. We show the difference to the ground truth for the L-MAE and the neural supervised-Att.}
  \label{fig:deformations}
\end{figure}

We evaluate this using local aneurysm deformations, shown in Figure~\ref{fig:deformations}. Source and target geometries differ mainly around the aneurysm. We mask this region, keep the source simulation outside the mask fixed, and inpaint the missing velocities. The quantitative results are shown in Table~\ref{tab:deformation}. Inside the edited mask, the supervised neural surrogate obtains an nMSE of $72.97\% \pm 9.81\%$, while L-MAE reduces this to $26.43\% \pm 4.19\%$. L-MAE also substantially improves directional agreement, increasing cosine similarity from $47.30\% \pm 4.48\%$ to $80.49\% \pm 3.60\%$, and reduces both divergence and vorticity error. On the full sample, the benefit is even more pronounced because the unchanged part of the simulation is preserved by construction: full-field nMSE decreases from $35.81\% \pm 2.97\%$ for the forward surrogate to $4.52\% \pm 0.83\%$ for L-MAE. In comparison, the FM models struggle in this task. We hypothesise that L-MAE's superior performance in local editing stems from its training objective, which directly optimises for conditional completion from partial tokens, whereas the global noise integration in latent flow-matching struggles to strictly adhere to narrow, high-frequency spatial boundary constraints. Grouping the edits by volume IoU shows that the inpainting error grows significantly with deformation magnitude for L-FM and L-MAE, giving an empirical estimate of the pipeline's limit, whereas supervised-Att shows no trend since it always predicts from scratch (Appendix~\ref{appendix:deformation_size}).

These results demonstrate a practical advantage of boundary-conditioned inpainting beyond full-field prediction. When local design or anatomical changes are made, \emph{our model can reuse the known global solution and focus its capacity on the region that actually changed}.

\begin{table}
\caption{Quantitative results for small geometry edits. We find that inpainting works especially well for only minor adaptations. We evaluate the masked region and the full simulation.}
\centering
\begin{tabular}{lcccc}
\toprule
Method & nMSE $\downarrow$ (\%) & CosSim $\uparrow$ (\%) & divergence $\downarrow$ & vorticity $\downarrow$ \\
\midrule
\multicolumn{5}{l}{\textit{Inside mask}} \\
Supervised-Att & 72.97\%$\,\pm\,$9.81\% & 47.30\%$\,\pm\,$4.48\% & 0.34$\,\pm\,$0.01 & 1.18$\,\pm\,$0.06 \\
L-FM-Zero & 89.25\%$\,\pm\,$19.69\% & 46.63\%$\,\pm\,$6.03\% & 0.24$\,\pm\,$0.01 & 1.22$\,\pm\,$0.12 \\
L-FM-Physics & 62.73\%$\,\pm\,$14.56\% & 57.94\%$\,\pm\,$5.87\% & \textbf{0.21$\,\pm\,$0.01} & 1.09$\,\pm\,$0.10 \\
L-MAE & \textbf{26.43\%$\,\pm\,$4.19\%} & \textbf{80.49\%$\,\pm\,$3.60\%} & 0.22$\,\pm\,$0.01 & \textbf{0.73$\,\pm\,$0.03} \\
\midrule
\multicolumn{5}{l}{\textit{Full sample}} \\
Supervised-Att & 35.81\%$\,\pm\,$2.97\% & 72.49\%$\,\pm\,$1.16\% & 0.29$\,\pm\,$0.01 & 0.97$\,\pm\,$0.02 \\
L-FM-Zero & 7.88\%$\,\pm\,$1.57\% & 88.35\%$\,\pm\,$1.27\% & 0.20$\,\pm\,$0.001 & 0.42$\,\pm\,$0.03 \\
L-FM-Physics & 7.14\%$\,\pm\,$1.55\% & 90.10\%$\,\pm\,$1.25\% & \textbf{0.19$\,\pm\,$0.01} & 0.42$\,\pm\,$0.03 \\
L-MAE & \textbf{4.52\%$\,\pm\,$0.83\%} & \textbf{94.41\%$\,\pm\,$0.72\%} & 0.19$\,\pm\,$0.01 & \textbf{0.34$\,\pm\,$0.02} \\
\bottomrule
\end{tabular}
\label{tab:deformation}
\end{table}

\section{Discussion}

We propose to view steady CFD inference as a context-conditioned inpainting problem rather than as a fully supervised forward prediction task. This changes the role of conditioning: boundary conditions, partial measurements, or unchanged regions from previous simulations are represented directly as observed flow context, rather than as task-specific input variables. Our experiments on intracranial aneurysm hemodynamics show both the promise and the current scope of this formulation. With only sparse inlet-outlet context, inpainting can recover full velocity fields, although a supervised forward surrogate remains highly competitive in the in-distribution setting for which it was explicitly trained. Inpainting is therefore not intended to replace boundary-conditioned surrogates in this classical setting, but to complement them when additional solution context is available, for example from a pre-treatment simulation for virtual interventions, 4D flow MRI or Doppler ultrasound. The advantage of inpainting becomes clearer when additional context is available and under distribution shift. In particular, the latent masked autoencoder is more robust to unseen mass flows and external dataset shift than the supervised baselines, and local geometry-editing experiments show that inpainting can reuse unchanged regions of an existing simulation instead of predicting the entire field from scratch. These results support the view that self-supervised inpainting can act as a reusable CFD prior, especially when the inference-time conditioning pattern differs from the training-time supervision used by conventional surrogates.

An adjacent, popular approach to neural surrogate modelling is autoregressive solver emulation (e.g., \cite{lippe2023pde}).
In contrast, our approach is end-to-end surrogate modelling (compare, e.g., \cite{alkin2024universal}).
Both approaches have their advantages and disadvantages: the former sacrifices broadness in scope (e.g., beyond a certain flow case) for close adherence to established solver behaviour, while the latter can be more generalisable.

\paragraph{Broader impact.} Inpainting physics could provide fast flow estimates across disciplines from medicine to aerospace, but since the models are not yet flawless, they should complement rather than replace numerical simulation.

\paragraph{Limitations.}
At the same time, our study is an initial step.
We focus on steady flows, mainly velocity fields in internal hemodynamics. Pressure, wall shear stress (WSS) and external aerodynamics are evaluated only in the first experiments in the Appendix. 
Derived biomedical or engineering metrics beyond WSS are not yet predicted.
Extending the formulation to transient flows, via autoregressive rollouts or spatio-temporal inpainting, is left for future work.
Moreover, the studied hemodynamic setting is relatively favourable for inpainting because inlet-outlet velocity context provides strong information about the dominant flow structure.
We also acknowledge that local context may contain dynamical parameters but may not contain structural parameters that determine the flow or simulation results, such as Reynolds number or choice of turbulence model.
More general CFD problems may require additional conditioning, particularly for properties that are difficult to infer from sparse context, such as viscosity or mesh resolution towards the domain boundary (e.g., artery wall).
Future work should test how the model handles physically inconsistent context, such as boundary velocities that violate continuity.

Overall, our results suggest that CFD inference need not always be formulated as a task-specific forward surrogate with a fixed set of input parameters. A self-supervised model trained to complete velocity fields can instead be conditioned at inference time on arbitrary known regions of the solution. This inpainting perspective provides a flexible alternative for boundary-condition shifts, partial observations, and local geometry edits, and points toward CFD models that behave less like single-purpose predictors and more like reusable generative priors over physically plausible flow fields, with possible implications for large-scale pre-training and foundation models.

\begin{ack}
This research was funded by the German Ministry of Research, Technology and Space (BMFTR) within the Software Campus program, funding code: 16IS23069. The responsibility for the content of this publication lies with the authors.
\end{ack}

\newpage
\bibliographystyle{plainnat}
\bibliography{bib}
\newpage
\appendix

\section{Metrics}

\label{appendix:metrics}
We validate the velocity field using direct metrics and additional metrics that assess the solution's orientation and physical plausibility. This way, we evaluate not only how well our generative model inpaints the velocity field, but also how strongly the solution adheres to the underlying physics.

Unless otherwise stated, we report the standard error over the test set in plots and tables. For specific implementation details, we refer to our code at \url{https://github.com/jonasw247/inpainting-physics}.

Let $\mathbf{x}_i \in \mathbb{R}^3$ be the positions of each point $i$ for a given sample, with
predicted velocity $\hat{\mathbf{v}}_i \in \mathbb{R}^3$ and ground-truth
velocity $\mathbf{v}_i \in \mathbb{R}^3$.

\paragraph{Normalised MSE (nMSE).} We report the sample-wise normalised error:
\begin{equation}
  \mathrm{nMSE}(\hat{\mathbf{v}}, \mathbf{v})
  \;=\;
  \frac{\sum_{i} \lVert \hat{\mathbf{v}}_i - \mathbf{v}_i \rVert_2^2}
       {\sum_{i} \lVert \mathbf{v}_i \rVert_2^2}
\end{equation}

\paragraph{Cosine similarity.}
To measure directional agreement independently of magnitude, we report the
mean cosine similarity between predicted and ground-truth velocity vectors:
\begin{equation}
  \mathrm{CosSim}(\hat{\mathbf{v}}, \mathbf{v})
  \;=\;
  \frac{1}{N}\sum_{i=1}^{N}
  \frac{\hat{\mathbf{v}}_i \cdot \mathbf{v}_i}
       {\lVert \hat{\mathbf{v}}_i \rVert_2 \, \lVert \mathbf{v}_i \rVert_2 }.
\end{equation}

We exclude the lowest 10\% of $\mathbf{v}_i$ in each sample. 

\paragraph{Normalised divergence.}
For an incompressible flow, the continuity equation requires
$\nabla \cdot \mathbf{v} = 0$. We estimate the velocity Jacobian
$J_i \in \mathbb{R}^{3\times 3}$ at each point from its $k$ nearest
neighbours by local least squares on the first-order Taylor expansion
$\mathbf{v}(\mathbf{x}_j) - \mathbf{v}(\mathbf{x}_i) \approx J_i^{\!\top} (\mathbf{x}_j - \mathbf{x}_i)$,
and report the divergence residual normalised by the average Jacobian
Frobenius norm so that the score is dimensionless and insensitive to the
overall flow scale:
\begin{equation}
  \mathrm{Div}(\hat{\mathbf{v}})
  \;=\;
  \frac{\tfrac{1}{N}\sum_{i=1}^{N} \lvert \operatorname{tr}(\hat{J}_i) \rvert}
       {\tfrac{1}{N}\sum_{i=1}^{N} \lVert \hat{J}_i \rVert_F + \varepsilon}.
\end{equation}
We use $k=10$ neighbors and subsample to at most $50{,}000$ query points per
sample.

\paragraph{Normalised vorticity RMSE.}
The vorticity $\boldsymbol{\omega} = \nabla \times \mathbf{v}$ is recovered
from the antisymmetric part of the Jacobian,
$\omega_i = (\partial_y v_z - \partial_z v_y,\; \partial_z v_x - \partial_x v_z,\; \partial_x v_y - \partial_y v_x)$.
We report the RMSE between predicted and ground-truth vorticity fields,
normalised by the mean ground-truth vorticity magnitude:
\begin{equation}
  \mathrm{Vort}(\hat{\mathbf{v}}, \mathbf{v})
  \;=\;
  \frac{\sqrt{\tfrac{1}{N}\sum_{i=1}^{N} \lVert \hat{\boldsymbol{\omega}}_i - \boldsymbol{\omega}_i \rVert_2^2}}
       {\tfrac{1}{N}\sum_{i=1}^{N} \lVert \boldsymbol{\omega}_i \rVert_2 + \varepsilon}.
\end{equation}

Both fields share the same KNN structure, so the comparison is not
biased by differing neighbourhoods.

\section{Model implementation details}
\label{appendix:model_details}

For the main training runs, we use an NVIDIA A100 GPU - 80GB and train each model for two days, selecting the best validation checkpoint. All models are set up to a similar size of 30 million parameters.

\paragraph{L-FM.}
A per-token DiT~\cite{peebles2023scalable} operates on the $2{,}500\times 256$ latent tokens produced by the NT. The backbone uses hidden size $512$, $8$ blocks, $8$ attention heads, and an MLP ratio of $4$. Token positions are embedded from the ball-centre coordinates; flow-matching time is injected through AdaLN-zero conditioning. We train with the rectified flow-matching objective and a logit-normal time schedule ($\mu{=}0,\sigma{=}1$). Optimisation uses AdamW with learning rate $10^{-4}$, weight decay $0.01$, gradient clipping at $1.0$, batch size $8$, and an EMA of $0.9999$ (warmup $1{,}000$ steps). 

\paragraph{L-MAE.}
A masked autoencoder is trained over the same NT latent tokens. The encoder and decoder are $4$-block transformers with hidden size $512$, $8$ heads and MLP ratio $4$. The decoder receives mask tokens for the held-out positions and is trained to regress the masked latents under MSE. To prevent the trivial position-prior shortcut we observed early on, the mask ratio is sampled per batch from $\mathcal{U}(0.1, 0.9)$. Optimisation uses AdamW with learning rate $4\times 10^{-4}$, weight decay $0.01$, gradient clipping at $1.0$ and batch size $24$.

\paragraph{Supervised-Att.}
We implement a hierarchical encoder over $5{,}000$ supernodes (constructed from $50{,}000$ subsampled mesh points via $k$-NN with $k{=}32$), a Perceiver-style geometry block, and a cross-attention decoder back to the query points, following~\cite{alkin2025ab}. The model has hidden dimension $384$, $6$ heads, $1$ geometry block and a \texttt{pscscscscs} encoder pattern with $2$ decoder blocks. Mass flow, wall distance and inlet distance are included through Fourier feature conditioning. We train with Lion ($\beta_1{=}0.9,\beta_2{=}0.99$, weight decay $0.05$), learning rate $2.5\times 10^{-5}$, gradient clipping at $0.25$ with $10$-epoch warmup and a cosine schedule, mixed-precision, plus per-sample SO(3) rotation and translation augmentation. Predictions on the $50{,}000$-point subset are upsampled to the full mesh via inverse-distance-weighted interpolation before metric evaluation.

\paragraph{Supervised-MP.}
A $5$-level feature-propagation architecture following~\cite{qi2017pointnet++}. Each level downsamples by a factor of $4$ ($\text{ratio}=0.25$) and queries up to $32$ neighbours. Feature propagation uses $k{=}4$ inverse-distance interpolation. The input is the $50{,}000$-point subsampled mesh and predictions are upsampled the same way as for the Supervised model. We train with AdamW (learning rate $4.24\times 10^{-4}$, weight decay $0.01$, gradient clipping at $1.0$), batch size $64$, cosine schedule with $5$-epoch warmup plus per-sample SO(3) rotation and translation augmentation.

\section{Neighbourhood tokeniser (NT)}
\label{Appendix:NT}

\begin{figure}[h]
  \centering
  \includegraphics[width=\textwidth]{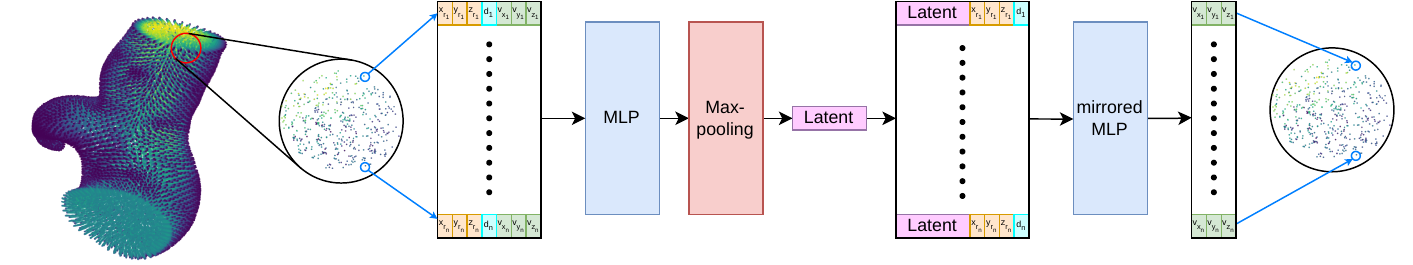}
  \caption{\textbf{Overview of the architecture of the neighbourhood tokeniser (NT).} (1) We obtain each neighbourhood and for every point we obtain the corresponding input for the encoder, containing $x_r, y_r, z_r$ as local position, $d$ as distance to the wall and $v_x, v_y, v_z$ for the velocity values. (2) We encode the data through an MLP followed by max-pooling, yielding the latent space for the neighbourhood. (3) We expand by creating copies of the latent space for each point in the neighbourhood and assign via concatenation the information on local position and distance to the wall. (4) Lastly, a similar mirrored MLP architecture predicts the velocity values on each point.}
  \label{fig:lonae}
\end{figure}

NT is based on a PointNet-style autoencoder \cite{qi2017pointnet} presented in Figure \ref{fig:lonae}. It consists of a shared MLP in the encoder and the decoder, and a max pooling layer to reduce the data into a latent space.
The MLP of the encoder consists of a concatenation of linear layers with ReLU activation functions and a layer normalisation right at the end of the encoder in order to ensure that the outcome of the MLP is normalised across channels. The MLP of the decoder is also a combination of linear layers and ReLU activation functions, but without a layer norm in comparison to the encoder. To account for the use of ReLUs, the weights of the hidden layers were initialised using He initialisation \cite{he2015delving}. The architecture of the designed NT, also performs a concatenation of the local position and distance to the wall after the latent space has been expanded to the same number of points required and before is fed into the decoder. This promotes the NT to focus only into the predictions of the velocity values.

The model is trained with MSE loss, Adam ($\alpha{=}1.6{\times}10^{-4}$), batch size 64, and a
ReduceLROnPlateau scheduler on training arteries.
SO(3) rotation augmentation is applied per artery per epoch, rotating both positions and velocities consistently.

The inputs of the autoencoder are the local position normalised, the distance to the wall and the velocity values, a 7-dimensional input. The local position is defined as the relative position to the centre of each of the neighbourhoods, which we normalise so every neighbourhood presents distances from zero (centre) to one (furthest point). The distance to the wall is considered here to be the Euclidean distance. The points in the wall are found by assuming the velocity to be zero near walls.

Moreover, the points that might have not been considered in any ball (large arteries) obtain their prediction based on interpolation from the 8 closest points that have got a prediction.

The number of centres has been evaluated by us. We evaluate the reconstruction quality under varying levels of compression (dictated by the number of centres and therefore local neighbourhoods). We utilize nMSE for the corresponding analysis under 548 arteries that have not been seen during training (the average of them is presented). Each model has been trained on the same arteries and splits for one day.

\begin{figure}[t]
  \centering
  \includegraphics[width=0.7\textwidth]{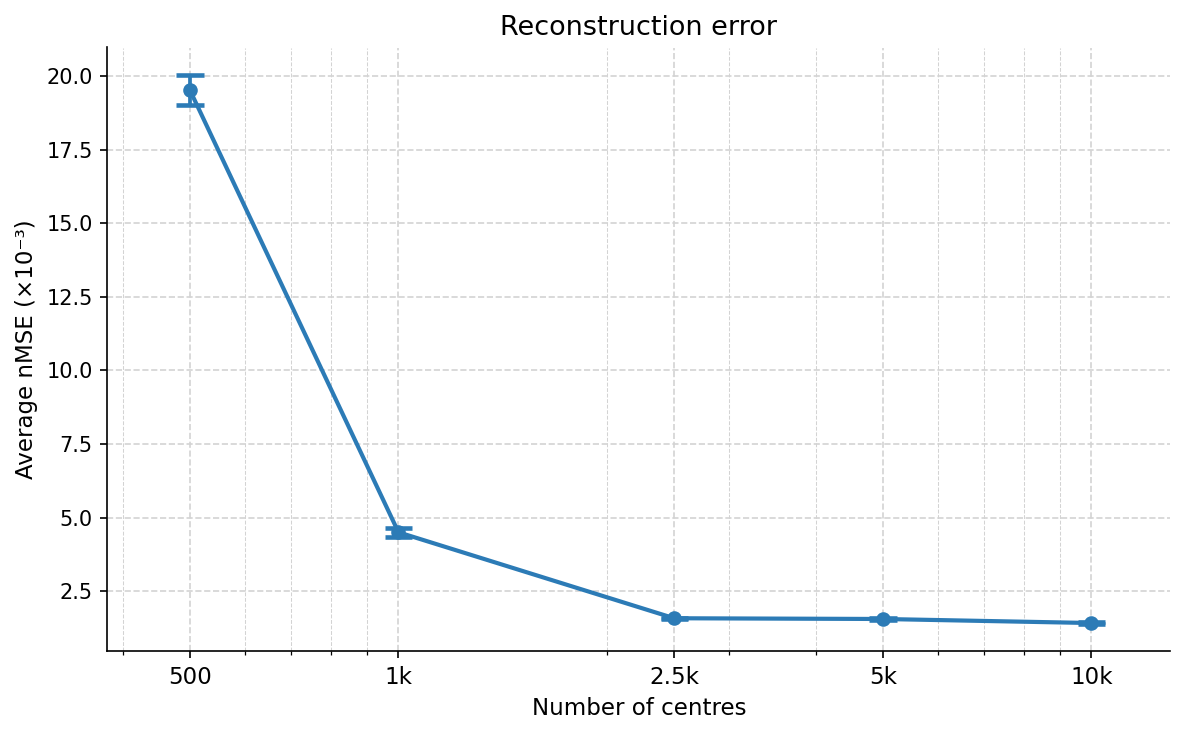}
  \caption{Analysis for nMSE values under different number of centres. 2500 is the best scenario providing the fewest possible number of centres without losing reconstruction capabilities.}
  \label{fig:centres}
\end{figure}

In Figure \ref{fig:centres} we observe that 2500 centers has almost the same nMSE compared to 10000, which allows us to reduce 4 times the number of centres while maintaining almost the same level on reconstruction quality.

The reason why the method establishes a constant number of points per neighbourhood area is for better resource management in terms of GPU. This occurs as by all having same dimensionality between batches, we do not require any masking criteria or alike. The number of points $N=512$ has been decided based on an Optuna study \cite{akiba2019optuna}.

Lastly, we also provide here ablation studies that demonstrate the results under the removal of key designs that have led to the final version presented via the NT.

The key decisions under analysis are:

\textbf{No neighbourhood normalisation:} We remove the normalisation of the relative positions to the centre of each ball, leading to the model having to consider different radii on balls as an influence factor.

\textbf{No data augmentation:} We remove the SO(3) rotations on each epoch.

\textbf{No distance to wall information:} We remove the scalar value dictating the distance to the wall as extra information in the input and also from the concatenation before the decoder stage.

\textbf{No local position and no distance to wall:} We remove the concatenation before the decoder part, by not adding any information on the position.

\textbf{Mean pooling:} We test mean pooling instead of max pooling.

We present the results in Table \ref{tab:ablation_nmse} for the listed experiments by using nMSE under the same split as we used for the number of centres and each model being trained for one day. We can observe that the final version provides the best results, as expected. Moreover, we can observe that the model can not reconstruct the data without the local position added before the decoder. In addition, the distance to the wall also represents a large improvement in terms of reconstruction results. The normalisation of balls and usage of max pooling also allow for better results. Finally, we observe that rotating the arteries as augmentation also provides an improvement.

\begin{table}[t]
  \caption{Average nMSE across different neighbourhood tokeniser variants. Final version is the one providing best results.}
  \label{tab:ablation_nmse}
  \centering
  \begin{tabular}{lc}
    \toprule
    \textbf{Variant} & \textbf{Average nMSE ($\times 10^{-3}$)} \\
    \midrule
    Final version                        & $1.68 \pm 0.03$ \\
    No augmentation                      & $1.74 \pm 0.03$ \\
    No normalisation                     & $4.54 \pm 0.07$ \\
    Mean pooling                         & $3.09 \pm 0.05$ \\
    No wall distance                     & $27.28 \pm 0.12$ \\
    No local pos. in decoder           & $185.50 \pm 0.93$ \\
    \bottomrule
  \end{tabular}
\end{table}

\section{Additional results}
\subsection{Additional OOD results}
\begin{figure}[h]
  \centering
  \includegraphics[width=\textwidth]{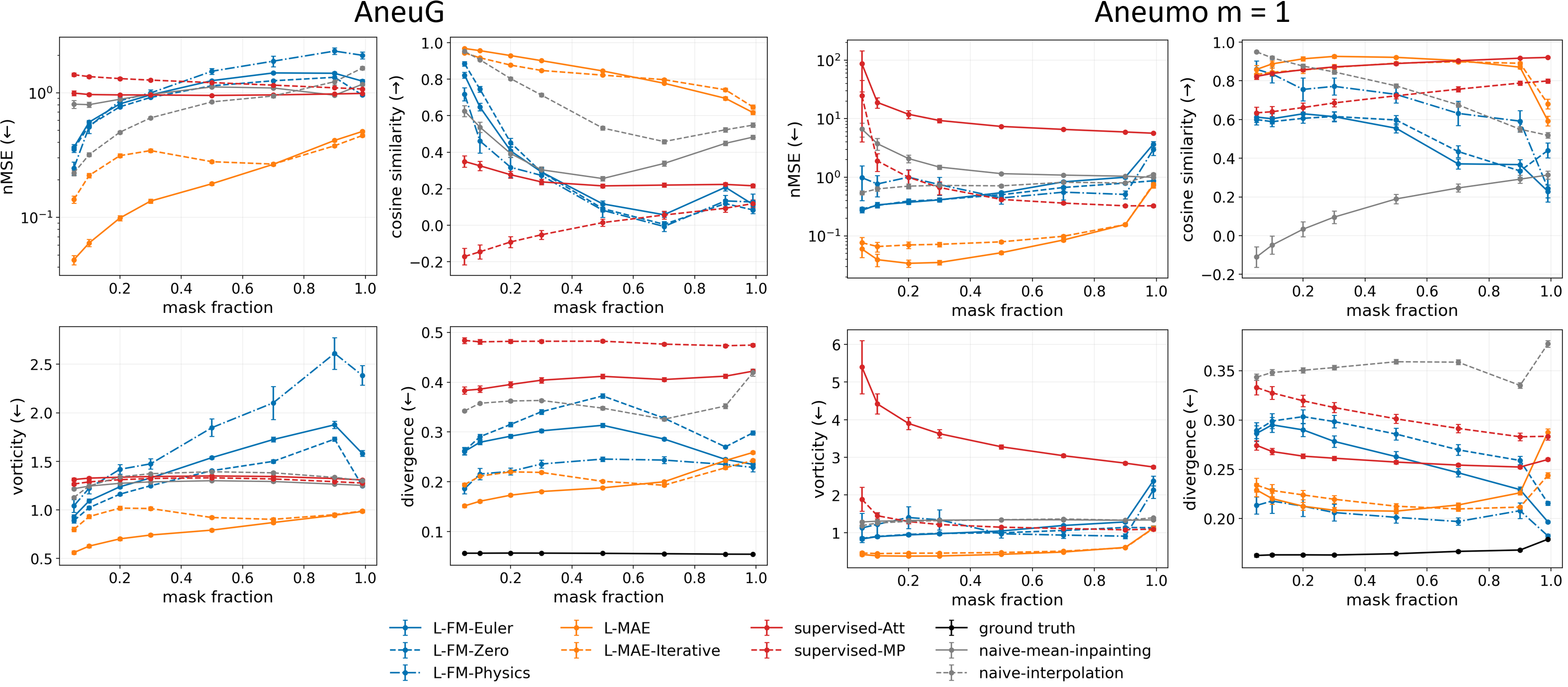}
  \caption{
  \textbf{The supervised models' prediction breaks for the external AneuG dataset and OOD mass flows. L-MAE inpainting is the only method consistently outperforming the naive baselines.} We compare different supervised and self-supervised models on the prediction of velocity fields provided with varying levels of context. Experiments are conducted on the external AneuG \cite{ding2025aneug} dataset. We show the respective results for Aneumo in Figure \ref{fig:mask_frac}.}
  \label{fig:mask_frac_AneuG}
\end{figure}

In Figure \ref{fig:mask_frac_AneuG}, we show the OOD results for the best methods on AneuG and on an OOD mass flow.

\subsection{Further flow matching integration schemes}
\begin{figure}[h]
  \centering
  \includegraphics[width=0.7\textwidth]{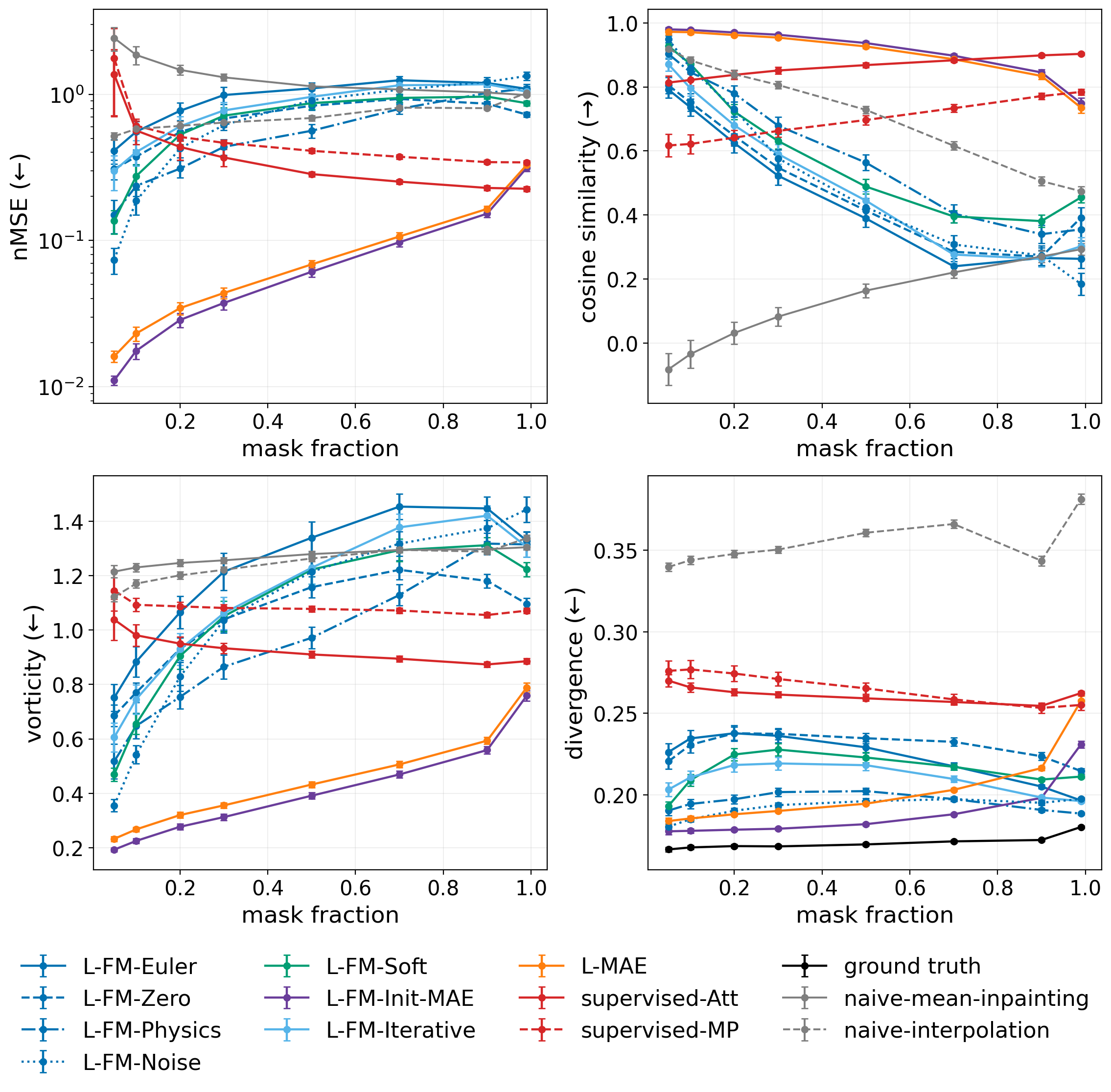}
  \caption{
  \textbf{Combining L-FM with L-MAE improves performance slightly.} We test additional integration schemes for L-FM. Iterative masking or soft boundaries show partial improvement, and noising the known tokens to the current time (L-FM-Noise) helps for small mask fractions, while initialising L-FM with the solution of the L-MAE slightly improves the L-MAE. We evaluate on Aneumo dataset on mass flow 3.}
  \label{fig:mask_frac_FM}
\end{figure}

In Figure \ref{fig:mask_frac_FM}, we test additional L-FM integration schemes:

\textbf{L-FM-Soft:}
During inpainting, we add soft boundaries by adding noise outside the mask.

\textbf{L-FM-Iterative:}
Similar to L-MAE-Iterative, we solve the inpainting problem in steps, starting from the outside and progressing toward the inside of the mask.

\textbf{L-FM-Init-MAE:}
We combine L-MAE prediction and refine it with FM integration. We add some noise ($t = 0.7$) to the L-MAE and integrate from there using L-FM.

\textbf{L-FM-Noise:}
Fixing the non-masked tokens to their clean latents creates a mixed-time latent state that the model has not seen during training. L-FM-Noise instead applies the correct noise level to all non-masked tokens, $z_t = (1-t)\,\epsilon + t\,z$, with a single fixed noise sample $\epsilon$. Compared to L-FM-Zero, it works better for small mask fractions ($50\%$ relative nMSE improvement at mask fraction $10\%$) and worse for large mask fractions ($41\%$ relative nMSE increase at $90\%$). This can be explained as follows: for small masked regions, a large portion of the latent is at the ``wrong'' time, whereas for large masked regions, the model mostly sees the correct time.

\subsection{Including pressure}
\begin{figure}[h]
  \centering
  \includegraphics[width=0.8\textwidth]{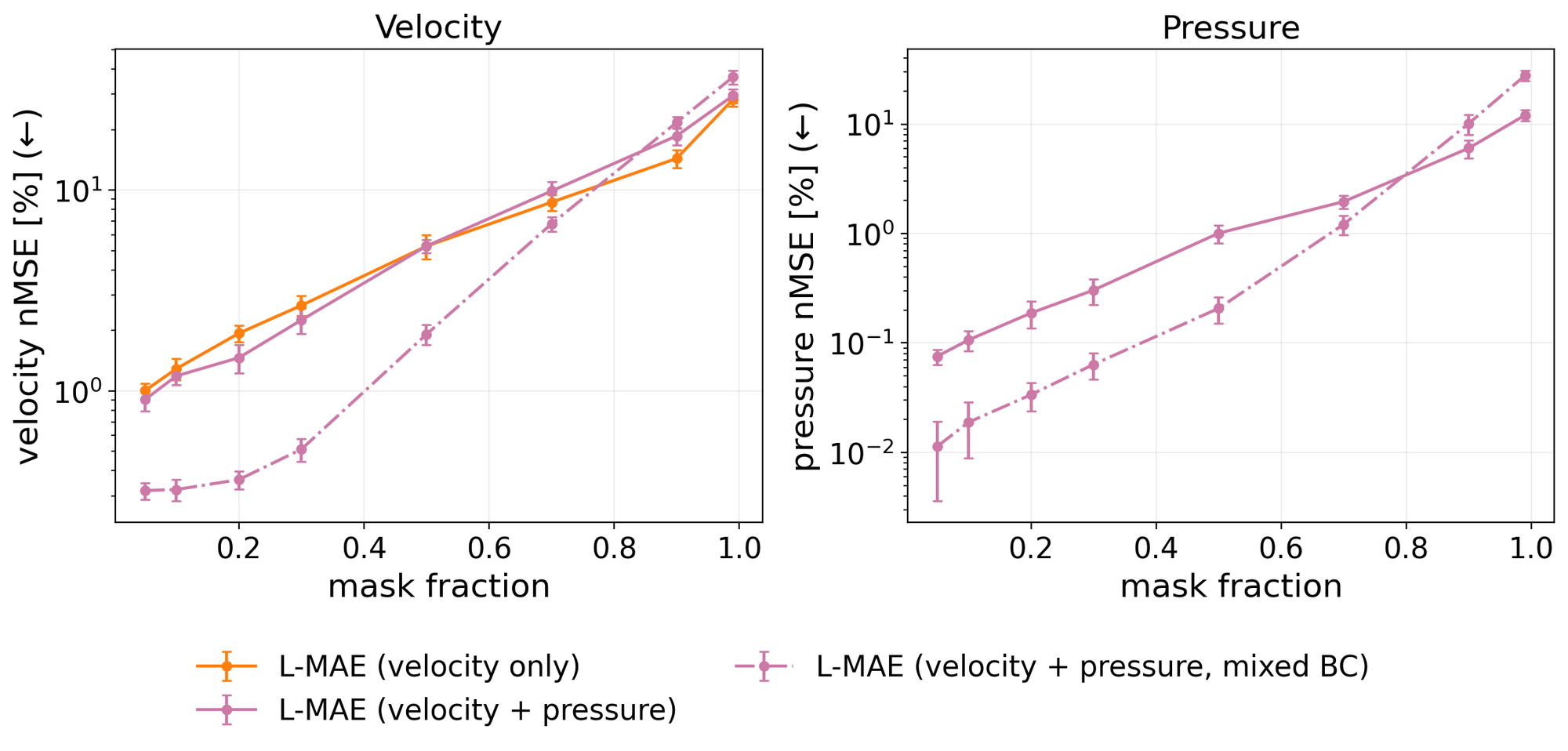}
  \caption{\textbf{Adding pressure as a second field preserves velocity accuracy, and mixed boundary conditions improve both fields when context is rich.} We train L-MAE jointly on velocity and pressure tokens and compare it to the velocity-only L-MAE. We show velocity (left) and pressure (right) nMSE over mask fraction for paired masking (the same region is masked in both fields) and mixed boundary conditions (velocity given only at the inlet, pressure only at the outlet). We evaluate on 50 Aneumo test arteries and report the median.}
  \label{fig:pressure}
\end{figure}

To showcase the ability to include multiple fields, we add pressure. 
We trained a separate neighbourhood tokeniser to encode the pressure field. We observed that some non-wall points in the Aneumo dataset had zero pressure and removed these points as artefacts. Using velocity and pressure tokens, 2,500 per field per sample, we train our best method (L-MAE). Using the combined L-MAE, we match velocity accuracy as shown in Figure \ref{fig:pressure}. For the pressure, nMSE drops below 11 \% for mask fractions under 0.9, which corresponds to a very sparse context.

With the pressure predictions available, we can compute the residual of the Navier-Stokes momentum equation and compare it to that of the ground-truth simulation. We observe an approximate 5-fold increase in momentum residuals over the ground truth for the sparse case with a 99 \% masking fraction. At a 50 \% masking fraction, we observe a momentum residual approximately twice as large as the ground truth. Given that this residual is highly sensitive to local errors in both velocity and pressure gradients, these values still indicate good physical consistency, particularly at moderate mask fractions.

Additionally, we evaluate the pressure for mixed boundary conditions, in which we fix the inlet velocity and outlet pressure. This deviation further improves velocity and pressure predictions for small mask fractions, but decreases performance at low context.

\subsection{Detailed analysis of geometric edits}
\label{appendix:deformation_size}

To investigate the relation between the strength of the local geometric edit and performance, we divide the deformation pairs of Table~\ref{tab:deformation} into three equal categories based on their volume intersection-over-union. Table~\ref{tab:deformation_size} reports the velocity nMSE per category and the Spearman correlation $\rho$ between deformation size and error (positive: larger edits give larger errors). As expected, we find no trend for the supervised case, as the prediction is run from scratch each time. L-FM and L-MAE, on the other hand, show a significant increase in error for larger deformations:

The full-sample numbers quantify the error in the complete domain. We additionally report the error inside the mask and outside the mask in isolation. Inside the mask, the L-MAE error increases with deformation size. Outside the mask, the error purely reflects reusing the source simulation for the edited target geometry. This error is small compared to the inpainting error within the mask. These results indicate that, across the tested deformation range, the dominant error arises within the edited region rather than from reusing the surrounding source simulation.

\begin{table}[ht]
\centering
\caption{\textbf{Inpainting error grows with the size of the geometric edit, mainly inside the edited region.} Velocity nMSE for small, medium, and large deformations, binned by volume IoU, on the full sample, inside the mask, and outside the mask. Mean $\pm$ standard deviation over the pairs in each bin; Spearman $\rho$ between deformation size and nMSE over all pairs.}
\label{tab:deformation_size}
\small
\begin{tabular}{lcccc}
\toprule
Method & Small & Medium & Large & Spearman $\rho$ \\
\midrule
\multicolumn{5}{l}{\textit{Full sample}} \\
L-FM-Zero & 4.5\%$\,\pm\,$4.3\% & 6.8\%$\,\pm\,$4.3\% & 12.7\%$\,\pm\,$12.7\% & $+0.51$ ($p=0.005$) \\
L-FM-Physics & 4.7\%$\,\pm\,$5.0\% & 5.3\%$\,\pm\,$3.9\% & 11.5\%$\,\pm\,$12.8\% & $+0.40$ ($p=0.034$) \\
L-MAE & \textbf{1.9\%$\,\pm\,$1.5\%} & \textbf{4.5\%$\,\pm\,$3.3\%} & \textbf{7.3\%$\,\pm\,$6.1\%} & $+0.59$ ($p=0.001$) \\
Supervised-Att & 44.3\%$\,\pm\,$21.4\% & 28.6\%$\,\pm\,$6.5\% & 30.6\%$\,\pm\,$8.8\% & $-0.26$ ($p=0.180$) \\
\midrule
\multicolumn{5}{l}{\textit{Inside mask}} \\
L-MAE & \textbf{19.7\%$\,\pm\,$11.1\%} & \textbf{24.7\%$\,\pm\,$8.3\%} & \textbf{35.8\%$\,\pm\,$36.7\%} & $+0.43$ ($p=0.021$) \\
Supervised-Att & 78.2\%$\,\pm\,$33.9\% & 73.1\%$\,\pm\,$64.4\% & 66.1\%$\,\pm\,$65.9\% & $-0.40$ ($p=0.033$) \\
\midrule
\multicolumn{5}{l}{\textit{Outside mask}} \\
L-MAE & \textbf{1.1\%$\,\pm\,$1.2\%} & \textbf{3.0\%$\,\pm\,$3.2\%} & \textbf{3.8\%$\,\pm\,$3.5\%} & $+0.48$ ($p=0.009$) \\
Supervised-Att & 43.1\%$\,\pm\,$20.5\% & 26.5\%$\,\pm\,$6.5\% & 28.8\%$\,\pm\,$8.1\% & $-0.32$ ($p=0.094$) \\
\bottomrule
\end{tabular}
\end{table}

\subsection{Generalisation to other flow domains}

\begin{figure}[h]
  \centering
  \includegraphics[width=0.8\textwidth]{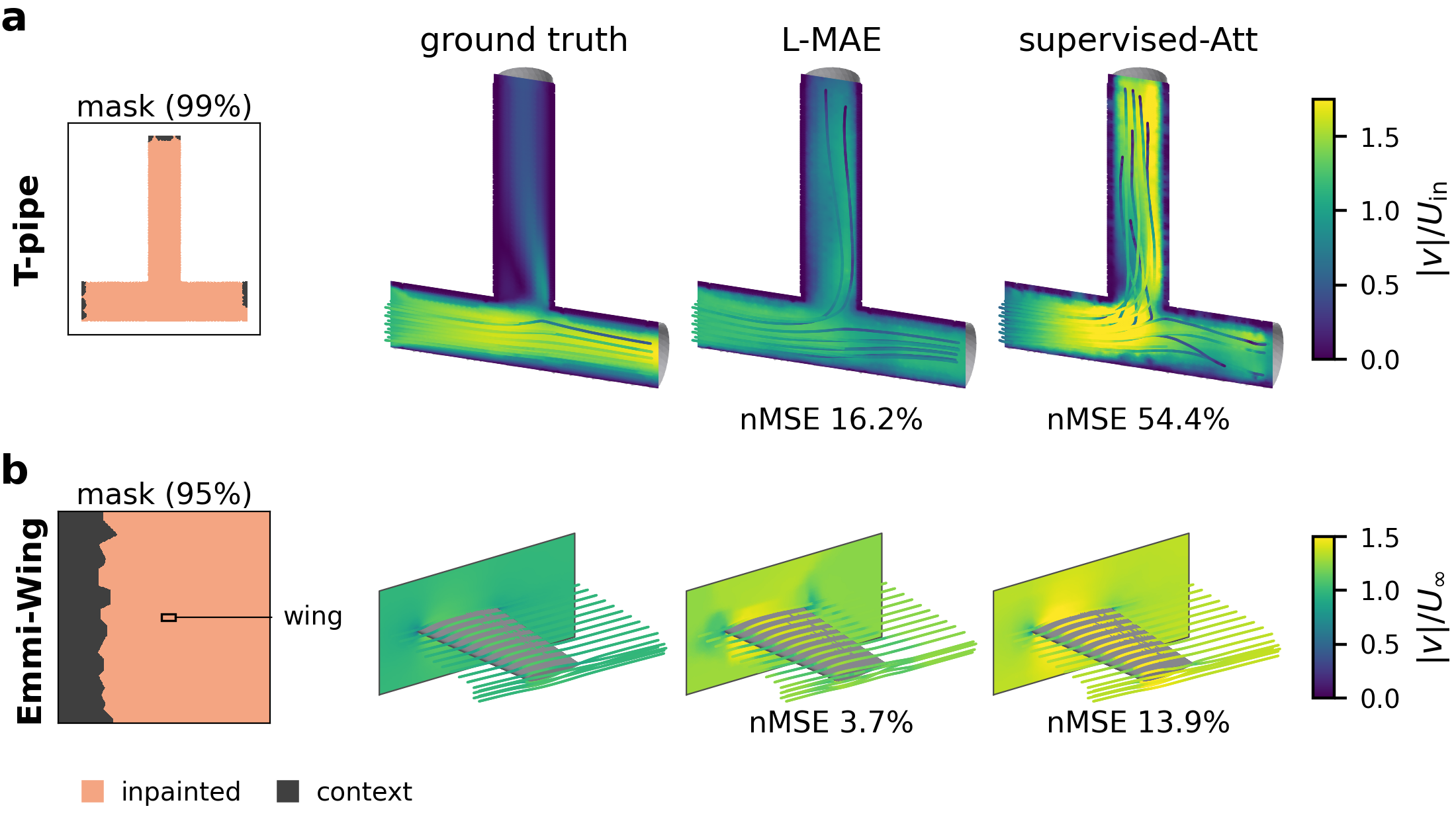}
  \caption{\textbf{L-MAE generalises to unseen flow domains and boundary conditions.} 
(a) Zero-shot T-pipe: models trained on Aneumo only, applied without retraining. Only the three openings are given as context (mask fraction 99\%). 
(b) Emmi-Wing at an OOD-slow inflow (153\,m/s): only the upstream free-stream is given as context (mask fraction 95\%). 
In both cases, L-MAE recovers the flow structure, whereas supervised-Att overestimates the velocity.}
  \label{fig:wing}
\end{figure}

\paragraph{Internal flow: T-pipe.}
To evaluate generalisation to an internal flow beyond arteries, we simulate a simple T-pipe geometry, where a pipe is connected orthogonally to the middle of another pipe, using the simpleFoam solver from OpenFOAM to generate a laminar, steady-state flow field. We directly apply the models trained on Aneumo, without any modification or retraining, to study generalisation to a widely out-of-domain problem. As context, we keep the tokens closest to the three openings (inlet and both outlets) and inpaint the rest (Figure~\ref{fig:wing}a). Sweeping over the mask fraction (as in Figure~\ref{fig:mask_frac}), L-MAE outperforms both supervised models by a large margin, whereas L-FM degrades at larger mask fractions (Table~\ref{tab:tpipe}).

\begin{table}
\caption{\textbf{Transfers zero-shot to a T-pipe, while L-FM and the supervised models degrade.} nMSE of the velocity field on an artery-sized T-pipe simulated with OpenFOAM, using the models trained on Aneumo without retraining. The context consists of the inlet and both outlets, and all methods are evaluated on the masked region. Results are for a single geometry.}
\centering
\begin{tabular}{lcccc}
\toprule
 & \multicolumn{4}{c}{Mask fraction} \\
\cmidrule(lr){2-5}
Method & 20\% $\downarrow$ & 50\% $\downarrow$ & 70\% $\downarrow$ & 99\% $\downarrow$ \\
\midrule
Supervised-Att & 90.6\% & 71.6\% & 69.2\% & 54.4\% \\
Supervised-MP & 76.5\% & 64.4\% & 59.5\% & 43.7\% \\
L-FM-Zero & 9.1\% & 42.7\% & 68.3\% & 141.2\% \\
L-MAE & \textbf{5.8\%} & \textbf{7.1\%} & \textbf{8.9\%} & \textbf{16.2\%} \\
\bottomrule
\end{tabular}
\label{tab:tpipe}
\end{table}
\paragraph{External flow: Emmi-Wing.}
To demonstrate that our method also generalises to other flow domains, we retrain it on external aerodynamics, the Emmi-Wing dataset \cite{paischer2025going}, a vastly different, turbulent regime from the arterial flows: external flow around wings at inflow velocities of $150$--$300\,$m/s instead of internal, low-velocity blood flow. Since the flow orientation is always the same, we removed rotation augmentation. We train our best method, the combination of tokeniser with L-MAE, and the state-of-the-art attention-based supervised model on a mid-range band of inflow velocities. As context, L-MAE only observes the free-stream inflow, given by the $5\%$ most upstream tokens in the far field, and inpaints the remaining $95\%$ of the domain, including the entire near-wing flow, while supervised-Att receives the inflow velocity and angle of attack explicitly (Figure~\ref{fig:wing}b). While performing worse than supervised-Att in the training regime (ID), L-MAE performs better out-of-domain (OOD) for velocities both slower and faster than the training velocities, confirming the superior generalisability of our method (Table~\ref{tab:wing}). This mirrors our findings for unseen mass flows on Aneumo (Figure~\ref{fig:surrogate_mass_flow}): a self-supervised flow prior conditioned on observed context extrapolates more robustly than a forward surrogate tied to its training-time conditioning range.

\begin{table}
\caption{\textbf{Generalisation to external aerodynamics on Emmi-Wing.} nMSE of the velocity field for models trained on a mid-range band of inflow velocities. The neighbourhood tokeniser, retrained on Emmi-Wing, reconstructs the velocity field with an nMSE of $0.04\%$ in-distribution and at most $0.11\%$ out-of-distribution, far below the inpainting error. We evaluate on wings inside the training band (ID), and on wings with slower and faster inflow (OOD).}
\centering
\begin{tabular}{lccc}
\toprule
Method & ID $\downarrow$ & OOD-slow $\downarrow$ & OOD-fast $\downarrow$ \\
\midrule
Supervised-Att & \textbf{0.45\%$\,\pm\,$0.16\%} & 13.80\%$\,\pm\,$0.27\% & 7.69\%$\,\pm\,$0.32\% \\
L-MAE & 1.46\%$\,\pm\,$0.34\% & \textbf{4.21\%$\,\pm\,$0.27\%} & \textbf{5.18\%$\,\pm\,$0.29\%} \\
\bottomrule
\end{tabular}
\label{tab:wing}
\end{table}

\subsection{Wall shear stress}
The risk of intracranial aneurysm rupture has been associated with wall shear stress (WSS)~\cite{zhou2017association}, making it an important downstream biomarker. We estimate WSS based on the predicted velocity field via the near-wall velocity gradient. L-MAE yields the lowest WSS error across all mask fractions (Table~\ref{tab:wss}), whereas L-FM and the supervised baselines perform considerably worse, indicating that L-MAE better preserves the near-wall velocity gradients required for WSS estimation.

\begin{table}[h]
\caption{\textbf{L-MAE best preserves the near-wall velocity gradients required for WSS estimation.} WSS nMSE derived from the predicted velocity field over mask fraction for supervised and self-supervised models. We evaluated 50 Aneumo test arteries.}
\centering
\small
\begin{tabular}{lccccc}
\toprule
Method & 20\% & 50\% & 70\% & 90\% & 99\% \\
\midrule
Supervised-Att & 213.4$\pm$34.0 & 149.3$\pm$14.6 & 129.3$\pm$14.2 & 121.9$\pm$14.4 & 83.9$\pm$7.3 \\
Supervised-MP & 125.8$\pm$28.0 & 72.6$\pm$6.4 & 66.6$\pm$6.3 & 66.3$\pm$6.7 & 51.2$\pm$3.7 \\
L-FM & 62.3$\pm$13.3 & 80.5$\pm$36.5 & 57.7$\pm$9.2 & 49.8$\pm$2.4 & 50.5$\pm$2.3 \\
L-MAE & \textbf{9.3$\pm$0.8} & \textbf{12.6$\pm$0.8} & \textbf{14.9$\pm$0.8} & \textbf{21.0$\pm$1.1} & \textbf{32.0$\pm$1.9} \\
\bottomrule
\end{tabular}
\label{tab:wss}
\end{table}

\subsection{Dataset licenses.} We use Aneumo~\citep{li2025aneumo} (CC BY-NC-ND 4.0), AneuG-Flow~\citep{ding2025aneug} (CC BY-SA 4.0) and Emmi-Wing~\citep{paischer2025going} (CC BY-NC 4.0) in accordance with their licenses, and generated the T-pipe reference simulation with OpenFOAM (GPL-3.0).

\end{document}